# CUDLE: Learning Under Label Scarcity to Detect Cannabis Use in Uncontrolled Environments

Reza Rahimi Azghan[1], Nicholas C. Glodosky[2], Ramesh Kumar Sah[2], Carrie Cuttler[2], Ryan McLaughlin[2], Michael J. Cleveland[2], Hassan Ghasemzadeh[1]

*Abstract*— Wearable sensor systems have demonstrated a great potential for real-time, objective monitoring of physiological health to support behavioral interventions. However, obtaining accurate labels in free-living environments remains difficult due to limited human supervision and the reliance on self-labeling by patients, making data collection and supervised learning particularly challenging. To address this issue, we introduce *CUDLE* (Cannabis Use Detection with Label Efficiency), a novel framework that leverages self-supervised learning with real-world wearable sensor data to tackle a pressing healthcare challenge: the automatic detection of cannabis consumption in free-living environments. CUDLE identifies cannabis consumption moments using sensor-derived data through a contrastive learning framework. It first learns robust representations via a self-supervised pretext task with data augmentation. These representations are then fine-tuned in a downstream task with a shallow classifier, enabling CUDLE to outperform traditional supervised methods, especially with limited labeled data. To evaluate our approach, we conducted a clinical study with 20 cannabis users, collecting over 500 hours of wearable sensor data alongside user-reported cannabis use moments through EMA (Ecological Momentary Assessment) methods. Our extensive analysis using the collected data shows that CUDLE achieves a higher accuracy of 73.4%, compared to 71.1% for the supervised approach, with the performance gap widening as the number of labels decreases. Notably, CUDLE not only surpasses the supervised model while using 75% less labels, but also reaches peak performance with far fewer subjects, indicating its efficiency in learning from both limited labels and data. These findings have significant implications for real-world applications, where data collection and annotation are labor-intensive, offering a path to more scalable and practical solutions in computational health.

## I. INTRODUCTION

Identifying instances of cannabis consumption is essential for various reasons. From a clinical standpoint, accurately detecting cannabis use is vital for understanding its impact on many physiological and psychological aspects, which helps in developing effective treatment protocols and interventions [1]. Moreover, real-time monitoring of cannabis consumption through mobile health sensor systems can help evaluate the effects of legalization policies and inform focused harm reduction strategies. Due to its legalization in many regions and its use for medical purposes, cannabis consumption has become increasingly prevalent in the United States. Stress is known to be a major contributing factor to addiction and chronic cannabis use is frequently used as a coping mechanism for stress [2]. As a result, stress-related biomarkers and signals collected by sensors can potentially help identify the moments of cannabis use.

This work was supported in part by the National Science Foundation, under grants CNS-2210133. Any opinions, findings, conclusions, or recommendations expressed in this material are those of the authors and do not necessarily reflect the views of the funding organizations.
[1]Arizona State University, Phoenix, AZ, USA
[2]Washington State University, Pullman, WA, USA
email: {rrahimia, hassan.ghasemzadeh}@asu.edu, {nicholas.glodosky, ramesh.sah, carrie.cuttler, ryan.mclaughlin, michael.cleveland}@wsu.edu

Mobile health technologies with sensors and machine learning hold great potential for cannabis detection in real-world settings [3] [4]. Intelligent machine learning algorithms deployed on ubiquitous mobile devices can provide unprecedented adoption of such technologies [5]–[7]. However, training a robust and generalized machine learning algorithm in the supervised learning paradigm requires extensive labeled data, and collecting labeled data using self-reported surveys in real-life conditions is a challenging and expensive process [8], [9]. Furthermore, sensor-based systems continuously collect signals without labels during operation, making the traditional supervised learning process unfeasible. In response, sophisticated techniques such as semi-supervised and self-supervised learning become suitable in these scenarios. Recent advancements in self-supervised learning for mobile health applications have used large-scale unlabeled data and achieved high performance [10]–[12]. Furthermore, self-supervised learning methods have demonstrated comparable or superior performance to supervised techniques for similar problems.

Self-supervised learning uses unlabeled data to learn the data characteristics by creating pretext tasks and training the model for detecting augmented data samples from original samples. The pretext task helps in learning a good representation of the data for the relevant downstream task. The main idea is to learn a higher level feature representation where the data itself provides supervision [13]. Self-supervised learning can be categorized into two main methods: generative and contrastive. Generative methods involve predicting part of the masked input, and contrastive methods implement data augmentation techniques to characterize similarities and differences between inputs. Contrastive learning involves training the model to encode similar samples in a way that their representations are also similar, while encoding dissimilar samples in a way that their representations are dissimilar. After contrastive training, the self-supervised block is frozen and connected to a trainable classification block to train for the original machine learning task using a few labeled data samples.

In this work, we introduce CUDLE, a self-supervised learning framework for cannabis detection using real-world sensor data. We have devised multiple data augmentation techniques within a contrastive loss-based self-supervised learning scheme, which we rigorously compared against traditional fully supervised methods. First part of our experiments target scenarios where all sensor data is available, but labels are only partially accessible, demonstrating that under these conditions, CUDLE significantly outperforms supervised learning approaches. Additionally, we tested the generalizability of both models by training each on data and labels collected from a limited number of participants. By progressively increasing the amount of training data, we evaluated how many subjects are necessary for each model to reach peak performance. Our findings indicate that CUDLE not only performs reliably with fewer labels but also generalizes effectively even with reduced sensor data

To the best of our knowledge, this work is the first to design a self-supervised learning approach for cannabis use detection. In short, contributions of this work can be summarized as follows:

- **Novel Application of CUDLE**: We present CUDLE, a self-supervised learning framework that uses contrastive loss and fine-tunes the downstream task with small amounts of labeled data. CUDLE is specifically applied to the detection of cannabis consumption using sensor-based Electrodermal Activity (EDA) data.
- **Label Efficiency with Sensors**: CUDLE surpasses traditional supervised learning models while utilizing 75% less labels, highlighting its ability to perform effectively with limited labels collected from wearable sensor devices.
- **Data Efficiency with Sensor Inputs**: Our findings reveal that CUDLE achieves peak performance with data from significantly fewer subjects compared to supervised models, highlighting its potential in sensor-driven applications where collecting large-scale labeled data is challenging.

## II. RELATED WORK

Using machine learning algorithms on sensor data to detect subject behaviors is a well-studied topic. In [14], [15], the authors utilized multiple sensor modalities to train a stress classification model. [16] and [17] focused exclusively on EDA data to monitor and detect stress. In [4], the authors employed a deep neural network to classify moments of cannabis use. While these studies applied sophisticated machine learning models to sensor data, the use of state-of-the-art methods, such as semi-supervised and self-supervised learning in uncontrolled environments, remains an underdeveloped area [18], [19].

Contrastive self-supervised learning was first used in computer vision applications to learn representation of images by contrasting positive pairs against negative pairs [20]. However, the idea of making representations of an input agree with its different transformation is much older [21]. Recently discriminative approach of self-supervised learning based on contrastive learning in the latent space have achieved impressive results in many applications [10]–[13], [22]. These approaches learns representation using objective function similar to loss function by training machine learning models to perform pretext tasks where both the inputs and labels are derived from an unlabeled data. In [22], the authors proposed *SimCLR* approach of contrastive self-supervised learning and showed significantly superior performance on image classification tasks compared to supervised learning methods. This led to numerous other related works on contrastive self-supervised learning in various application domains.

In [12], the authors extended the ideas presented in [22] for univariate time domain application with specific changes in the architecture of the learning model and data augmentation techniques to better fit the temporal dependencies of the time series data. They combined the dynamic time warping based pretext tasks capable of retaining time series structure with deep learning based time series classification method *InceptionTime* [23] to extract meaningful representation for the downstream time series classification tasks. Another extension of *SimCLR* for multi-variate time series application was presented in [13]. In [13], authors discussed novel data augmentation methods or pretext tasks and compared the performance of self-supervised method with supervised approach for fault classification at multiple labelling percentage.

In general, self-supervised methods based on contrastive learning has shown great potential in limited label conditions. However, the reason behind the success of constrastive methods is not clear yet. Some works have attributed the success of contrastive self-supervised learning to maximization of mutual information between latent representation [24] and other works have argued that the success is due to the specific form of the contrastive loss function [25].

## III. KEY COMPONENTS OF CUDLE

CUDLE comprises three key components: an initial data augmentation method specifically designed for time-series data, a pretext encoder that extracts the most relevant features from the entire sensor dataset, and a classifier trained on the (potentially limited) labeled data to generate the final output. Figure 1 illustrates the overall structure of CUDLE's components. We briefly discuss each of the key components in detail.

### A. Data Augmentation

Data augmentation refers to artificially increasing the size and diversity of the data by applying various transformations.

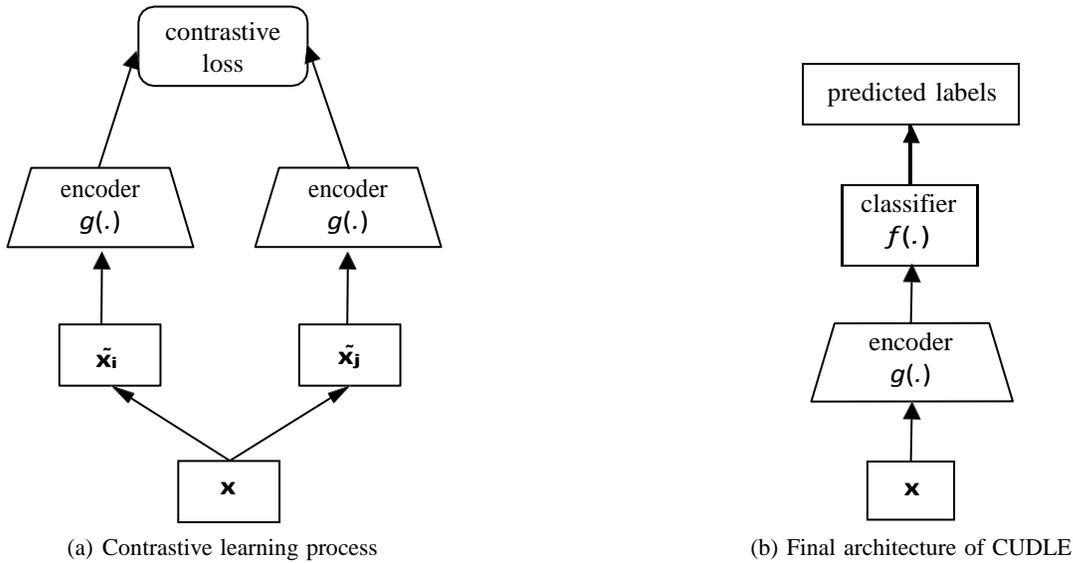

Fig. 1: The architecture of CUDLE as a whole. The left figure shows how data augmentation is utilized in contrastive learning to train the encoder component. The right figure illustrates how the trained encoder and classifier integrate to form the complete CUDLE system.

Data augmentation introduces different views of an instance into the subsequent encoder model. In other words, it helps the model pay attention to the most essential features in the structure of the data. Moreover, it enhances model generalizability, thereby increasing its effectiveness in differentiating between similar and dissimilar instances.

Augmentations in this study are applied to time series data, $D$, defined as below.

$$D = \{x_0^T, x_1^T, ..., x_{m-1}^T\}$$

Here $x_i$ represents the segments of the time series data, each belonging to $R^T$, where $T$ denotes the segment length. Inspired by [12], in this study, four kinds of data augmentation were implemented.

*1) Left-to-right-flipping:* Data obtained in this technique is the rotated-around-the-y-axis of the original data. This technique adds variability to the dataset by flipping the order of the data points without changing the signal's inherent characteristics. It theoretically should strengthen the capability of the following encoder model to generalize over patterns and will increase its power to become resilient to temporal changes. Mathematically, this augmentation can be expressed as follows:

$$\tilde{D} = DA$$

Where $D$ and $\tilde{D}$ are the original data and the augmented data, respectively, and both are in $R^{m \times T}$. $A$ is a permutation matrix, in $R^{T \times T}$ with entries defined as

$$A_{ij} = \begin{cases} 1 & i + j = T \\ 0 & \text{otherwise} \end{cases}$$

*2) Blockout:* This augmentation randomly takes a fraction of the signal and resets all its points to zero. In this case, the encoder model is going to learn how to predict and recognize the patterns, even with missing or unclear data points. The encoder, when trained on imperfect data this way, should be more robust and generalizing. Such an augmentation can be formally defined as follows:

$$\tilde{D} = DI_\lambda$$

Where $I_\lambda$ is the matrix with zeros off the main diagonal and almost all ones in the main diagnoal, expect for $\lambda \times T$ entries. In our implementation, $\lambda$ is set to $0.1$.

*3) Crop and resize:* In this method, we randomly select a continuous chunk of the time series segment and crop it. In our implementation, we take half of the given segment to be the size of this chunk. After cropping the sub-segment, we resize it back to the size of the original time series by interpolation. The interpolation method we use is averaging two successive values of the sub-segment.

In summary, this data augmentation technique crops a continuous chunk of length $T/2$ from the time series segment and then interpolates it back to restore the original length $T$. So, while basic patterns from the original data are captured, some variability is injected such that the encoder model generalizes much better with different patterns from the data.

*4) Random Gaussian noise:* : By adding small amounts of noise to each of the data points within the time series segment, this approach converts the original data into a slightly perturbed version, The placement and severity of the added perturbations are controlled by the mean ($\mu$) and a standard deviation ($\sigma$) of the noise. This augmentation technique allows the encoder model to become robust to small changes and noise in data and helps generalize over new data. Mathematically, this augmentation can be defined as:

$$\tilde{D} = D + D_{pert}$$

**Algorithm 1:** Crop and resize augmentation

**Input:** Original time-series segment $\mathbf{x} = [x_1, ..., x_T]$
**Result:** Augmented segment $\tilde{\mathbf{x}}$

Select a random starting point $i$ such that $0 \leq i < \frac{T}{2}$;
$\mathbf{x}_{crop} \leftarrow [\,]$;
**for** $k$ in range($\frac{T}{2}$) **do**
 $\mathbf{x}_{crop}[k] \leftarrow \mathbf{x}[i+k]$;
**end**
$\tilde{\mathbf{x}} \leftarrow [\,]$;
**for** $k$ in range(T) **do**
 $j \leftarrow \lfloor k/2 \rfloor$;
 $\tilde{\mathbf{x}}[k] \leftarrow \mathbf{x}_{crop}[j] + (k \mod 2) \times \frac{\mathbf{x}_{crop}[j+1] - \mathbf{x}_{crop}[j]}{2}$;
**end**
return $\tilde{\mathbf{x}}$

Where each row in $D_{pert}$ is a noise vector drawn from the Gaussian distribution $N(\mu, \sigma)$.

### B. Encoding

After applying the same augmentation to the original data twice, the augmented versions are fed into the encoder network, which generates two distinct embeddings. This encoder model is then trained with a contrastive loss function, aiming to reduce the distance between embeddings of similar instances (positive pairs) while increasing the distance between embeddings of different instances (negative pairs). This training process ensures that the model learns to distinguish between similar and dissimilar data effectively.

Two distinct representations, $\tilde{\mathbf{x}}_i$ and $\tilde{\mathbf{x}}_j$, which are considered as positive representations, are generated after applying two augmentations to the segment $\mathbf{x}$. Therefore, with a batch size of $N$, these augmentations result in 2 positive representations and $2N - 2$ negative representations. Once the encoder model $g(.)$ processes the entire batch, the similarity between the positive and the negative representations are calculated as [1]

$$l_{i,j} = -\log \frac{\exp(\text{sim}(\mathbf{z}_i, \mathbf{z}_j))}{\sum_{k=1}^{2N} \mathbb{1}_{k \neq i} \exp(\text{sim}(\mathbf{z}_i, \mathbf{z}_k))}$$

Where $\text{sim}(.,.)$ is the dot product of two vectors, $\mathbb{1}_{statement}$ returns one when the statement is true and return zero otherwise, and $\mathbf{z}_i = g(\tilde{\mathbf{x}}_i)$. The loss function is subsequently calculated for all possible positive pairs for every segment $\mathbf{x}$.

### C. Classification

A fully-supervised classification model that utilizes the available labels is designed and appended to the trained encoder of CUDLE. The first operation is to encode the labeled data to obtain the corresponding representations. The obtained representations, along with the associated class labels, are passed into the classifier to map the representations to their corresponding correct labels. A well-trained self-supervised encoder contains most of the underlying structure of the data. Therefore, in such an architecture, a classification network with few parameters can potentially give satisfactory performance.

[1]The temperature parameter $\tau$ is always set to 0.05 in this work.

The pre-processing and training procedure for both the encoder part and the classifier part is given in Algorithm 2.

**Algorithm 2:** CUDLE's training procedure

**Input:** dataset $D \in R^{m \times T}$, (partial) label vector $\mathbf{y} \in R_m$
**Result:** contrastive loss $l_{cl}$, cross-entropy loss $l_{ce}$
 Trained encoder $g_\vartheta$ and classifier $f_\theta$ models

Break $D$ into two parts, labeled segments $D_l$ and unlabeled segments $D_{ul}$;
**for** epoch in range $n_1$ **do**
 **for** batch in $D_{ul}$ **do**
  cl(batch) $\leftarrow$ 0;
  **for** all $\mathbf{x}$s in the batch **do**
   $\tilde{\mathbf{x}}_i, \tilde{\mathbf{x}}_j \leftarrow$ augment($\mathbf{x}$);
   $\mathbf{z}_i, \mathbf{z}_j \leftarrow g_\vartheta(\tilde{\mathbf{x}}_i), g_\vartheta(\tilde{\mathbf{x}}_j)$;
   cl(batch) $\leftarrow$ cl(batch) + $l(\tilde{\mathbf{x}}_i, \tilde{\mathbf{x}}_j,$ batch)
  **end**
  $\vartheta \leftarrow \vartheta - \eta \nabla_\vartheta$ cl(batch)
 **end**
**end**
**for** epoch in range $n_2$ **do**
 **for** batch in $D_l$ **do**
  $y \leftarrow$ label(batch);
  $\tilde{y} \leftarrow f_\theta(g_\vartheta(\text{batch}))$;
  $\theta \leftarrow \theta - \eta \nabla_\theta l_{ce}(\tilde{y}, y)$
 **end**
**end**

## IV. EXPERIMENTS

In this section, we begin by detailing our dataset specifics and outline the details of CUDLE's architecture, including its training configuration and structure. Following this, we detail the evaluation metrics and the process of evaluation. next, we analyze the performance of the different augmentation methods by comparing their effects on CUDLE. Finally, we demonstrate the results from the best augmentation method across different levels of data and labeling strength and compare them with the results we got with a fully-supervised model.

### A. Dataset and Model Settings

*1) Data collection in Free-living environment:* The initial step in our study was to collect data over a 24-hour period as subjects go about their daily routines. Our study included 20 healthy adult cannabis users from a small town in Washington state, recruited through various advertisements.

Participants completed an online screening survey to assess eligibility, which required them to own a smartphone, be at least 21 years old, fluent in English, and free of neurological, intellectual, or serious psychiatric conditions. They were excluded if they reported heavy recent illicit drug use, or use of corticosteroid medications. Eligible cannabis users had to use cannabis at least four times a week for at least a year. A research assistant delivered an Empatica E4 wristband (shown in figure 2) and instructions to participants' homes. Participants wore the wristband for 24 hours, recording continuous, real-time physiological indicators of sympathetic neural activity via measures of heart rate variability and electrodermal activity. Subjects also kept a log of cannabis use times, which later was used as labels for training and validating the machine learning algorithms [26].

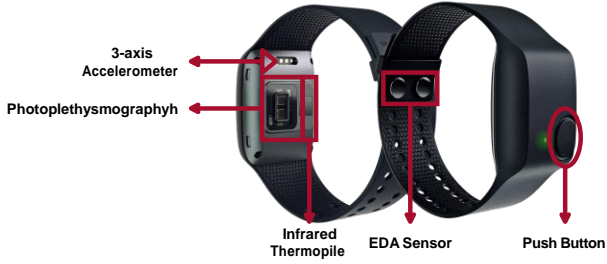

Fig. 2: Empatica E4 wristbands with embedded sensors

*2) Data Preprocessing:* An overlapping window algorithm was implemented to divide the EDA signal collected through the wristbands into segments of 1 hour long each, where every segment was overlapped by the previous one for 50 minutes [27]. Then, every segment was labeled as True in case participants used cannabis at any time instance of the segment, or False otherwise. The resulting dataset was a tabular data, where each row represented the raw EDA values of a segment, and the row's label corresponded to the segment's label.

$$D = \{x_0^\top, x_1^\top, ..., x_{m-1}^\top\}, \quad y = [y_0, ..., y_{m-1}]^\top$$

Where $x_i$s are in $R^T$ and $y$ is in $R^m$, $T$ is the length of the segment, and $m$ is the number of rows (segments) in the overall data.

Finally, a per-subject standardization was applied to the dataset to remove any inter-subject variability. The final dataset had the properties mentioned below

$$\forall s \subset S, \quad E_i[x_{i,s}] = 0, \quad \sigma_i(x_{i,s}) = 1$$

Where $S$ is the set of all the subjects, $E_i[x_{i,s}]$ and $\sigma_i(x_{i,s})$ are the average and the standard deviation of the segment values corresponding to subject $s$, respectively.

Following preprocessing with sliding window segmentation and subject-specific normalization, a dataset of 875 samples from 20 distinct cannabis users was created. Of these samples, 409 were labeled as true, indicating that the corresponding window overlapped with a cannabis intake event.

*3) Model Settings:* We implemented both a Convolutional Neural Network (CNN) [28] and a Multilayer Perceptron (MLP) [29] as potential encoders for CUDLE and compared the results generated by each. Both encoders were trained using a contrastive loss function, optimized with Adam, with a learning rate of 0.001 for 100 epochs. The training process included four different data augmentation techniques applied to the sensor-derived EDA signals: left-to-right flipping, block-out, Gaussian noise, and crop-and-resize. The effects of these augmentations on the EDA signals are illustrated in Figure 3.

Once the encoder was trained, a shallow MLP was attached as the classifier. The classifier was then trained using algorithm 2, minimizing cross-entropy loss over 100 epochs, with a learning rate of 0.001. During classifier training, the encoder parameters were frozen to preserve the learned sensor-based representations developed by CUDLE.

### B. Evaluation Process

In this study, four performance metrics are used to report the outcomes of CUDLE. We report accuracy, as the ratio of samples correctly-labeled by the model to the total number of samples. Precision is used to measure the model ability in avoiding errors when identifying the negative class (non-cannabis use). Recall is used to measure the model's effectiveness in detecting all positive instances (cannabis use). Finally, we report the F1-score, which is the weighted average of precision and recall.

To ensure a reliable evaluation of CUDLE, we used the leave-one-subject-out method [30] to avoid data leakage from training to testing data. Therefore, for each iteration, data segments of one subject were put aside as the test set, while data segments from all other subjects were used as the training set. This continued iteratively until each subject's data could be tested exactly once. Averaging of reported metrics across such train-test configurations ensures a robust performance. The reported metrics represent the average values obtained across all these train-test configurations.

### C. Experimental Results

Table I shows the outcomes from applying the four different data augmentation techniques while training CUDLE's encoders. The results were achieved by first training the encoder with these augmentations, then freezing the encoder's parameters, and finally training the classifier using all the available labels.

As evident from Table I, across all data augmentation methods, using an MLP model as the encoder improved the overall performance. This could be due to the MLP's ability to effectively capture non-linear relationships in the relatively low-dimensional EDA data. Additionally, the crop and resize augmentation technique delivered the best results with both models. This outcome is intuitively understandable: this method is similar to linearly interpolating the data and filling in missing values within the signal. Consequently, it preserves the overall structure of the signal while enriching the data with additional valuable information. This augmentation technique enables the encoder model to extract more meaningful patterns from the sensor-derived EDA signals. As a result, CUDLE utilizes an MLP as its encoder, and is trained with the crop-and-resize augmentation technique.

Figure 4 compares the results of CUDLE with those of a baseline model. Both CUDLE and the baseline model share the same architecture, but trained in different ways. The baseline model is also the encoder and the classifier model connected sequentially, but trained in a fully supervised manner without the help of data augmentation techniques.

To approximate a real-world scenario where labeling is labor-intensive and data annotations are often limited, we conduct the experiment with varying proportions of available labels. Specifically, we assume that the ratio $|D_{train,labeled}| / |D_{train}|$ is not always one, reflecting the reality that not all training data is fully labeled. We assume that labels are available from one, two, five, ten, and fifteen subjects, corresponding to 5%, 10%, 25%, 50%, and 75% of the data being labeled, respectively. It is important to note that the encoder part is still being trained on the entirety of $D_{train}$. On the contrary, the classifier model appended to the encoder, as well as the baseline model, are trained with varying levels of labeled data.

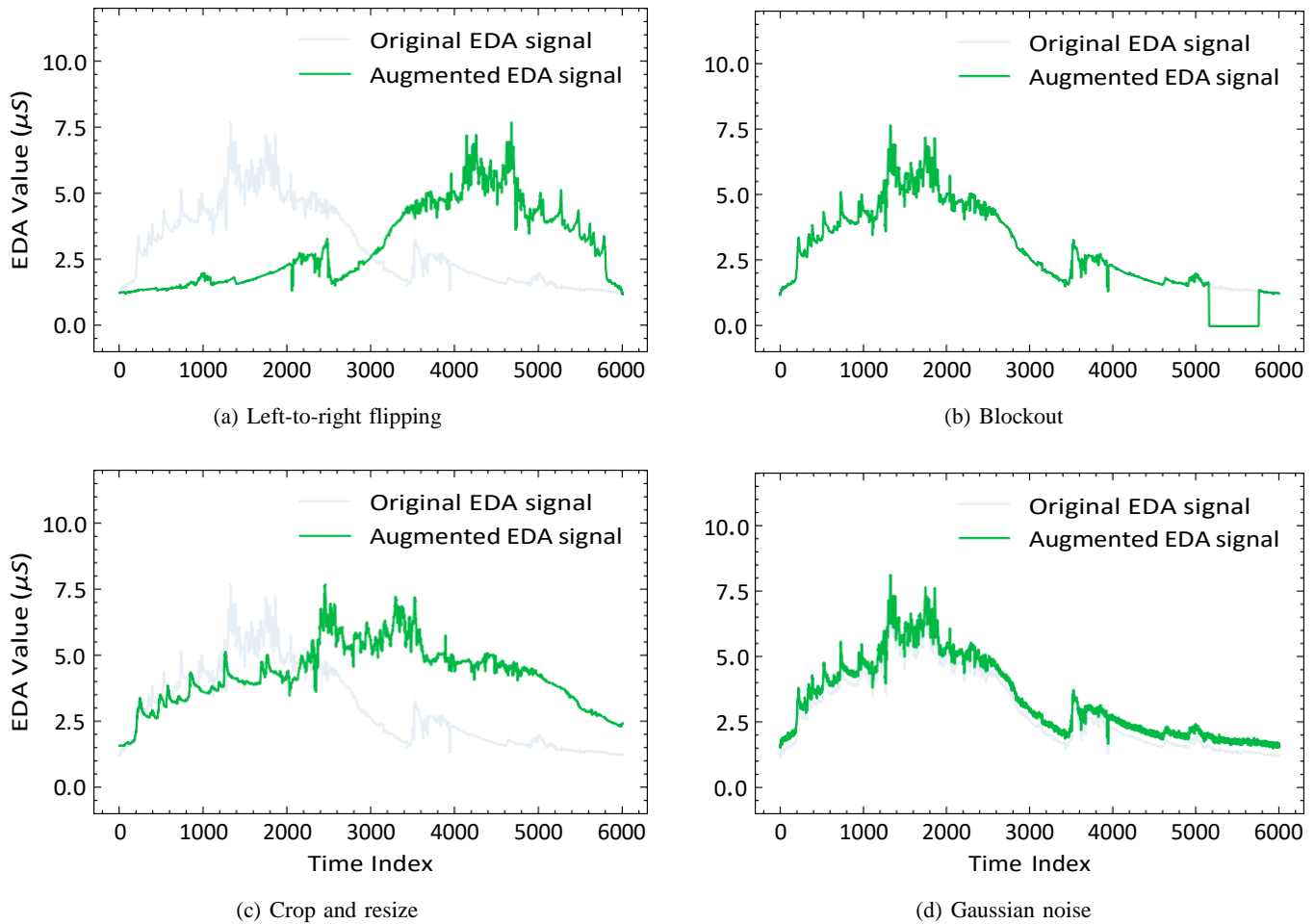

Fig. 3: Impact of four distinct augmentations on a random segment of the EDA signal.

TABLE I: Average performance measures of different augmentations and encoders on the test data

| Augmentation method | Encoder Model | Accuracy | Precision | Recall | F1-Score |
|---|---|---|---|---|---|
| Left-to-right flipping | CNN | 67.35 | 66.04 | 61.71 | 63.81 |
| | MLP | 70.21 | 71.18 | 65.43 | 68.17 |
| Blockout | CNN | 69.62 | 66.10 | 58.52 | 62.08 |
| | MLP | 71.43 | 71.24 | 62.12 | 66.37 |
| Crop-and-resize | CNN | 69.51 | 67.54 | 60.19 | 63.66 |
| | MLP | **73.55** | **73.92** | **66.03** | **69.75** |
| Random noise | CNN | 63.10 | 61.82 | 56.98 | 59.31 |
| | MLP | 66.67 | 68.38 | 58.33 | 62.97 |

Figure 4 demonstrates that CUDLE consistently outperforms the baseline model across all labeled data settings. From the robust performance of CUDLE, it can be concluded that the model can use even more unlabeled data to learn better. Another key advantage of CUDLE, as evident from the figure, is that it can obtain almost optimal performance even with a minimal labeled data. The results show that even if the data annotator provides only 25% of the labels, CUDLE still achieves its peak performance. This efficiency in learning from limited labeled data is particularly valuable in real-world scenarios where annotating data can be time-consuming and expensive.

Finally, we examine the generalizability of CUDLE by evaluating how many train subjects are needed for each model to reach its peak performance. In this experiment, we began by selecting one subject's data to serve as the test set. We then trained both CUDLE and the baseline model, starting with the data from a single subject and gradually increasing to data from 19 subjects. After each training iteration, we evaluated the trained model on the test data. We ensured that the training and test data came from different subjects in each iteration. This process was repeated until every participant's data had been used exactly once as the test set. The average results for each training configuration (ranging from one to 19 subjects) are presented in Figure 5. The experiment was conducted on our home institution's supercomputer, utilizing NVIDIA A100 tensor core GPUs.

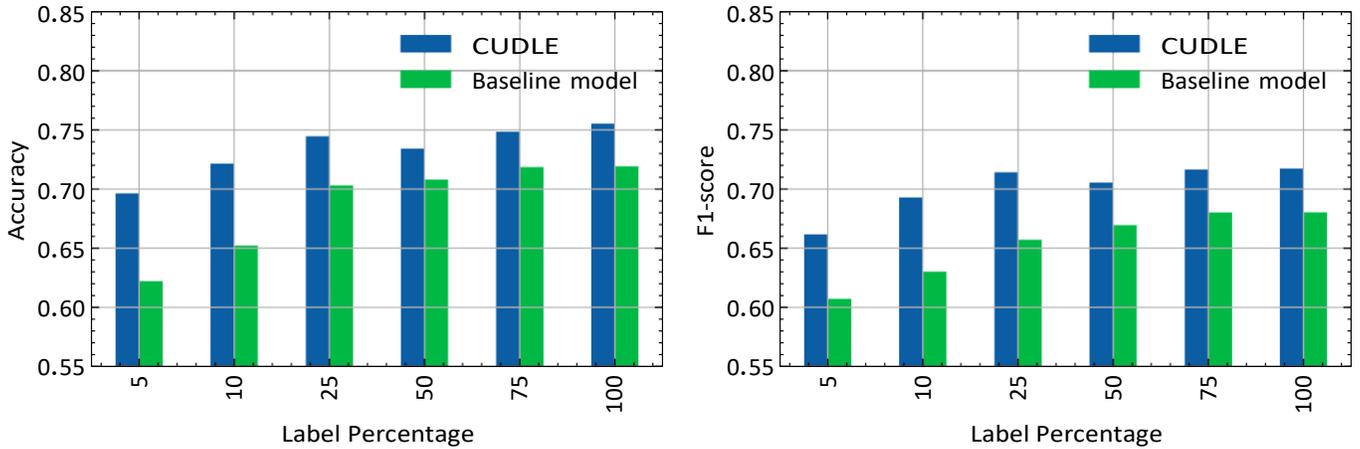

Fig. 4: Comparison of CUDLE and the baseline model under varying levels of labeling strength. As the number of labels increases, the performance of both models converges. However, with a lower percentage of available labels, the performance gap between the two models becomes more pronounced.

Figure 5 indicate that while both models eventually achieve similar peak performance, CUDLE reaches its peak with fewer subjects compared to the baseline model. This demonstrates CUDLE's robustness and ability to generalize effectively from limited data, as well as limited labels.

## V. CONCLUSION

In this study, we explored the use of electrodermal activity (EDA) data to detect cannabis consumption times through both supervised and self-supervised learning techniques. Our primary objective was to enhance model performance, particularly in scenarios with limited labeled data, by leveraging the strengths of self-supervised learning. Notably, the strong performance of the self-supervised learning approach across different levels of labeling strength highlights its practical value in clinical environments where obtaining labeled data is both challenging and costly. This work demonstrates significant potential for real-time monitoring of cannabis consumption, which could be instrumental in enhancing patient care through timely behavioral interventions and better management of conditions such as chronic pain or anxiety.

Looking ahead, several avenues for future research emerge from this study. Exploring additional data augmentation techniques, such as time warping, jittering, and synthetic data generation, could further enhance the variability and robustness of the training data, potentially leading to even better model performance. Moreover, integrating multimodal data, including heart rate, temperature, and accelerometer data, with EDA signals could provide a more comprehensive understanding of the user's physiological state.

## REFERENCES

[1] N. Boumparis and M. P. Schaub, "Recent advances in digital health interventions for substance use disorders," *Current Opinion in Psychiatry*, vol. 35, no. 4, pp. 246–251, 2022.
[2] J. D. Buckner, M. J. Zvolensky, A. H. Ecker, and E. R. Jeffries, "Cannabis craving in response to laboratory-induced social stress among racially diverse cannabis users: The impact of social anxiety disorder," *Journal of Psychopharmacology*, vol. 30, no. 4, pp. 363–369, 2016.
[3] S. Bae, T. Chung, B. Suffoletto, M. R. Islam, J. Du, S. Jang, Y. Nishiyama, R. Mulukutla, and A. K. Dey, "Mobile phone sensor-based detection of subjective cannabis "high" in young adults: A feasibility study in real-world settings," *Abstracts from the 2020 Virtual Scientific Meeting of the Research Society on Marijuana July 24th, 2020*, 2021.
[4] R. R. Azghan, N. C. Glodosky, R. K. Sah, C. Cuttler, R. McLaughlin, M. J. Cleveland, and H. Ghasemzadeh, "Personalized modeling and detection of moments of cannabis use in free-living environments," *2023 IEEE 19th International Conference on Body Sensor Networks (BSN)*, pp. 1–4, 2023.
[5] A. Mamun, S. I. Mirzadeh, and H. Ghasemzadeh, "Designing deep neural networks robust to sensor failure in mobile health environments," in *2022 44th Annual International Conference of the IEEE Engineering in Medicine & Biology Society (EMBC)*, 2022, pp. 2442–2446.
[6] S. M. Alsadat, J.-R. Gaglione, D. Neider, U. Topcu, and Z. Xu, "Using large language models to automate and expedite reinforcement learning with reward machine," 2024. [Online]. Available: https://arxiv.org/abs/2402.07069
[7] I. R. Rahman, S. B. Soumma, and F. B. Ashraf, "Machine learning approaches to metastasis bladder and secondary pulmonary cancer classification using gene expression data," in *2022 25th International Conference on Computer and Information Technology (ICCIT)*, 2022, pp. 430–435.

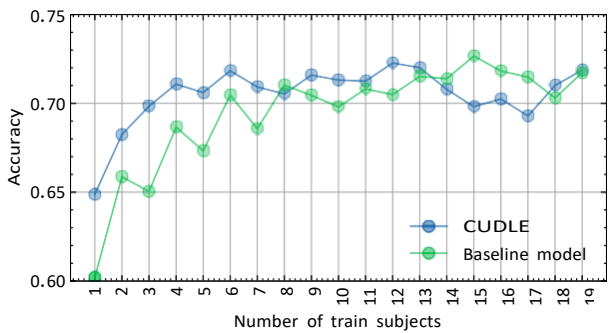

Fig. 5: Comparison of CUDLE and the baseline model across varying numbers of training subjects.

[8] R. Fallahzadeh and H. Ghasemzadeh, "Personalization without user interruption: Boosting activity recognition in new subjects using unlabeled data," in *Proceedings of the 8th International Conference on Cyber-Physical Systems*, 2017, pp. 293–302.

[9] Y. Ma and H. Ghasemzadeh, "Labelforest: Non-parametric semi-supervised learning for activity recognition," in *Proceedings of the AAAI Conference on Artificial Intelligence*, vol. 33, no. 01, 2019, pp. 4520–4527.

[10] S. Rabbani and N. M. Khan, "Contrastive self-supervised learning for stress detection from ecg data," *Bioengineering*, vol. 9, 2022.

[11] H. Yuan, S. Chan, A. P. Creagh, C. Tong, D. A. Clifton, and A. Doherty, "Self-supervised learning for human activity recognition using 700,000 person-days of wearable data," *NPJ Digital Medicine*, vol. 7, 2022.

[12] X. Yang, Z. Zhang, and R. yi Cui, "Timeclr: A self-supervised contrastive learning framework for univariate time series representation," *Knowl. Based Syst.*, vol. 245, p. 108606, 2022.

[13] J. Pöppelbaum, G. S. Chadha, and A. Schwung, "Contrastive learning based self-supervised time-series analysis," *Appl. Soft Comput.*, vol. 117, p. 108397, 2022.

[14] P. Schmidt, A. Reiss, R. Duerichen, C. Marberger, and K. Van Laerhoven, "Introducing wesad, a multimodal dataset for wearable stress and affect detection," in *Proceedings of the 20th ACM International Conference on Multimodal Interaction*, ser. ICMI '18. New York, NY, USA: Association for Computing Machinery, 2018, p. 400–408. [Online]. Available: https://doi.org/10.1145/3242969.3242985

[15] E. Farahmand, S. Sheikhpour, A. Mahani, and N. Taheri, "Load balanced energy-aware genetic algorithm clustering in wireless sensor networks," in *2016 1st Conference on Swarm Intelligence and Evolutionary Computation (CSIEC)*, 2016, pp. 119–124.

[16] R. K. Sah and H. Ghasemzadeh, "Stress classification and personalization: Getting the most out of the least," 2021. [Online]. Available: https://arxiv.org/abs/2107.05666

[17] S. A. H. Aqajari, E. K. Naeini, M. A. Mehrabadi, S. Labbaf, A. M. Rahmani, and N. Dutt, "Gsr analysis for stress: Development and validation of an open source tool for noisy naturalistic gsr data," 2020. [Online]. Available: https://arxiv.org/abs/2005.01834

[18] R. Holder, R. K. Sah, M. Cleveland, and H. Ghasemzadeh, "Comparing the predictability of sensor modalities to detect stress from wearable sensor data," in *2022 IEEE 19th Annual Consumer Communications & Networking Conference (CCNC)*, 2022, pp. 557–562.

[19] F. Elhambakhsh and M. Saidi-Mehrabad, "Developing a method for modeling and monitoring of dynamic networks using latent variables," *International Journal of Industrial Engineering and Production Research*, vol. 32, pp. 29–36, 03 2021.

[20] R. Hadsell, S. Chopra, and Y. LeCun, "Dimensionality reduction by learning an invariant mapping," in *2006 IEEE computer society conference on computer vision and pattern recognition (CVPR'06)*, vol. 2. IEEE, 2006, pp. 1735–1742.

[21] S. Becker and G. E. Hinton, "Self-organizing neural network that discovers surfaces in random-dot stereograms," *Nature*, vol. 355, no. 6356, pp. 161–163, 1992.

[22] T. Chen, S. Kornblith, M. Norouzi, and G. Hinton, "A simple framework for contrastive learning of visual representations," in *International conference on machine learning*. PMLR, 2020, pp. 1597–1607.

[23] H. Ismail Fawaz, B. Lucas, G. Forestier, C. Pelletier, D. F. Schmidt, J. Weber, G. I. Webb, L. Idoumghar, P.-A. Muller, and F. Petitjean, "Inceptiontime: Finding alexnet for time series classification," *Data Mining and Knowledge Discovery*, vol. 34, no. 6, pp. 1936–1962, 2020.

[24] P. Bachman, R. D. Hjelm, and W. Buchwalter, "Learning representations by maximizing mutual information across views," *Advances in neural information processing systems*, vol. 32, 2019.

[25] M. Tschannen, J. Djolonga, P. K. Rubenstein, S. Gelly, and M. Lucic, "On mutual information maximization for representation learning," *arXiv preprint arXiv:1907.13625*, 2019.

[26] P. Alinia, R. K. Sah, M. McDonell, P. Pendry, S. Parent, H. Ghasemzadeh, and M. J. Cleveland, "Associations between physiological signals captured using wearable sensors and self-reported outcomes among adults in alcohol use disorder recovery: development and usability study," *JMIR Formative Research*, vol. 5, no. 7, p. e27891, 2021.

[27] R. K. Sah, M. J. Cleveland, A. Habibi, and H. Ghasemzadeh, "Stressalyzer: Convolutional neural network framework for personalized stress classification," in *2022 44th Annual International Conference of the IEEE Engineering in Medicine & Biology Society (EMBC)*, 2022, pp. 4658–4663.

[28] Y. LeCun, Y. Bengio, and G. Hinton, "Deep learning," *nature*, vol. 521, no. 7553, pp. 436–444, 2015.

[29] S. Haykin, *Neural networks: a comprehensive foundation*. Prentice Hall PTR, 1994.

[30] T. Hastie, R. Tibshirani, J. H. Friedman, and J. H. Friedman, *The elements of statistical learning: data mining, inference, and prediction*. Springer, 2009, vol. 2.